\documentclass{article}
\usepackage{arxiv}

\usepackage[utf8]{inputenc}
\usepackage[T1]{fontenc}
\usepackage{microtype}

\usepackage{amsmath,amsfonts,amssymb}
\usepackage{graphicx}
\usepackage{booktabs}
\usepackage{algorithm}
\usepackage[noend]{algpseudocode}
\usepackage[numbers,sort&compress]{natbib}
\usepackage{hyperref}
\usepackage[capitalise,noabbrev]{cleveref}
\usepackage{orcidlink}

\newcommand{\dropcap}[1]{#1}
\newcommand{\siref}[1]{\cref{#1}}
\newcommand{\eqbreak}{}

\DeclareMathOperator{\pa}{\mathrm{Pa}}
\DeclareMathOperator{\de}{\mathrm{De}}

\title{Amortized Bayesian Inference on Multilevel Models of Arbitrary Structure}
\date{September 30, 2026}

\author{%
  Daniel Habermann\,\orcidlink{0000-0003-3685-7287} \\
  Department of Statistics \\
  TU Dortmund University, Germany \\
  \texttt{daniel.habermann@tu-dortmund.de} \\
  \And
  Andreas Bulling\,\orcidlink{0000-0001-6317-7303} \\
  Institute for Visualisation and Interactive Systems \\
  University of Stuttgart, Germany \\
  \And
  Stefan T. Radev\,\orcidlink{0000-0002-6702-9559} \\
  Department of Cognitive Science \\
  Rensselaer Polytechnic Institute, USA \\
  \And
  Paul-Christian Bürkner\,\orcidlink{0000-0001-5765-8995} \\
  Department of Statistics \\
  TU Dortmund University, Germany \\
  \texttt{paul.buerkner@gmail.com} \\
}

\begin{document}
\maketitle

\begin{abstract}
    We develop a general method for amortized Bayesian inference on multilevel models of arbitrary structure. Given a generative model specified as a directed acyclic graph, our method automatically derives valid factorizations of the joint posterior and matching neural network architectures. The key steps, graph expansion and graph inversion, yield an inverse graph that determines how inference networks are stacked and conditioned, producing factorizations that amortize over the number of groups and the number of observations within each group.
    Unlike approaches that simplify the dependency structure to speed up learning or inference, our method preserves all conditional independence and exchangeability assumptions of the generative model. Across three case studies, it closely matches gold-standard samplers on models with more than 6,500 parameters while reducing inference to a near-instant forward pass once trained.
\end{abstract}


\section{Introduction}\label{sec:introduction}

\dropcap{A}mortized Bayesian inference (ABI) approximates the posterior distribution of a Bayesian model by training neural networks on simulated pairs of parameters and datasets \citep{ZammitMangion2025}. Because it only requires simulation and no likelihood evaluation, ABI enables Bayesian inference for models that are otherwise computationally intractable. In contrast to classical approaches, such as approximate Bayesian computation \citep[ABC;][]{Marin2012}, it scales much better to high-dimensional data \citep{Dingeldein2025, Zhou2025, Orozco2025} and is also \textit{amortized}: the computational cost is only incurred once during training. This enables Bayesian inference for real-time applications \citep{Dax2025} as well as inference across millions of independent datasets \citep{Krause2022}. Consequently, ABI has become instrumental in a growing range of problems in the quantitative sciences \citep{Goncalves2020,Radev2021,Arruda2026,Brehmer2020,Wehenkel2023,Zhang2023}.

Multilevel models, however, are particularly challenging to amortize. The dimension of their posterior depends on the number of groups and can vary from one dataset to the next, which is at odds with the fixed-width inputs and outputs assumed by standard neural approximators. 

Recent neural network architectures relax the fixed-dimension requirement. Transformer-based approximators represent parameters and data as tokens and use masked attention to encode assumed dependencies, allowing arbitrary conditioning patterns \citep{Gloeckler2024}. Meta-amortized approaches generalize further, amortizing across families of structurally similar models with varying parameterizations, design matrices, and sample sizes \citep{Huang2026}. Neither addresses the dependency structure of multilevel models. Attention masks must be pre-specified for each model and do not guaranty conditional independencies, and meta-amortized architectures have so far been restricted to exchangeable observations, with exchangeable parameters left as an open problem.

Prior work \citep{Rodrigues2021,Arruda2026,Heinrich2024,Habermann2025,Kucharsky2026} has made progress on amortized inference for two-level models, deriving the network architecture by hand for that model class in each case. Deriving such an architecture is implicitly a choice of a particular factorization of the posterior: because each network has fixed input and output dimensions, a posterior whose dimension grows with the number of groups cannot be approximated in one piece but only as a product of factors that are learned separately and reused across groups. The choice of posterior factorization decides whether the approximator amortizes over the number of groups at all.
Two questions therefore arise: Does such a factorization exist for a given generative model, and if so, which factors are amortizable?

The defining assumption of a multilevel model is that parameters at a given level are exchangeable: they are drawn from a common distribution, whose hyperparameters are themselves estimated from the data. This assumption allows partial pooling of information across groups, rather than fitting each group in isolation or ignoring the grouping structure entirely \citep{Gelman2006}.
It also enables amortization: once we condition on the hyperparameters, a single neural network with shared weights can infer each group independently from its group-specific data alone \citep{Kucharsky2026, Habermann2025}. This requires a factorization of the joint posterior that makes the corresponding conditional independencies explicit, and not every factorization does. Even for a minimal two-level model, five valid factorizations exist (see \cref{sec:appendix-inverse-factorizations}), of which only two enable amortization over the number of groups.

For multilevel models of arbitrary structure and complexity, we show that a suitable factorization of the posterior can be determined from the model's \emph{generative graph} alone. It represents the model as an annotated directed acyclic graph (DAG) from which data can be simulated. To derive this factorization, we first apply \emph{graph expansion}, which splits each exchangeable node into two instances and thereby makes the exchangeability of groups explicit in the graph structure. \emph{Graph inversion} then enumerates the valid inverse factorizations of the expanded graph, which allows us to identify those that amortize over the number of groups. 
Even when no such factorization exists, we recover amortization by selecting a factorization that leaves the fewest group-level parameters conditioned on other groups. Then, we infer the remaining parameters autoregressively.

In contrast to prior work, in which the network architecture is derived by hand for a particular model class, our method determines the complete approximator automatically: which posterior factorization to use, which inference and summary networks are required, and how they are stacked and conditioned. Even though our focus lies on multilevel models, this extends amortized Bayesian inference to any model that can be expressed as a DAG, including crossed designs, for which no amortized method has previously been available.

We demonstrate our method in three case studies of increasing structural complexity: a two-level model of the eight schools data, a three-level model of inter-rater agreement on skin lesion segmentations with crossed image and annotator effects, and a four-level model of the UK Breeding Bird Survey with more than 6,500 parameters. In all three, the derived approximators closely match the posteriors obtained by Stan.

An implementation of our method is available in the open-source BayesFlow library \citep{Kuehmichel2026} as the \texttt{GraphicalApproximator} class.
\section{Methods} \label{sec:methods}

The core of our method is a four-step transformation of the generative graph into a complete neural network architecture for ABI. First, \emph{graph expansion} makes the exchangeability of groups explicit in the graph structure (\cref{sec:exchangeable-nodes}). Second, \emph{graph inversion} enumerates the valid factorizations of the posterior implied by the expanded graph (\cref{sec:graph-inversion,sec:graph-inversion-expanded}). Third, \emph{amortization analysis} compares the factorizations and selects the one that makes the most group-level parameters independently amortizable (\cref{sec:non-independent-amortization}). Fourth, \emph{architecture derivation} turns the selected inverse graph into a concrete set of approximators (\cref{sec:summary-networks,sec:network-architecture}): which nodes can share an inference network, how the networks are conditioned on one another, and how the observations are summarized for each of them.

\subsection{Generative graphs and ancestral sampling}\label{sec:graphical-simulator}

Almost all Bayesian models are generative, which means that they not only enable inference of parameter values $\theta$ conditioned on some observed data $y$, but also allow for generating \emph{simulated} data conditioned on $\theta$. This property is used extensively when following a principled Bayesian workflow \citep{Gelman2020}, for example, in the form of posterior predictive checks \citep{Gabry2019} by comparing actually observed data to data simulated under the assumptions of a model, or when performing prior elicitation \citep{Mikkola2024} by investigating the relationship between different prior specifications and hypothetical outcomes.

Let $G=(V,E)$ denote a DAG with a set of nodes $V$ and edges $E$, in which each node $v \in V$ represents a set of random quantities, and each edge encodes a conditional dependency. The nodes partition into latent nodes, whose values form the parameters $\theta$, and observed nodes, whose values form the data $y$. We refer to a latent node that is not a root as an \emph{interior} node. Because $G$ is a DAG, the joint distribution factorizes as
\begin{equation}\label{eqn:joint-factorization-general}
    p(\theta, y) = \prod_{v \in V} p( v \mid \pa_G(v)),
\end{equation}
where $\pa_G(v)$ denotes the parents of $v$ in $G$, and nodes with $\pa_G(v) = \emptyset$ are the root nodes, which are specified by their marginal prior distributions. Simulating new data can then be achieved by an ancestral sampling scheme: nodes are visited in topological order of $G$, starting from the roots and proceeding until the leaf nodes are reached. 

For illustration purposes, we will use a simple two-level model as a running example. We apply the same method to more complex three- and four-level models in \cref{sec:case-studies}. The example model underlies the eight schools study \citep{Rubin1981}, a meta-analysis of coaching effects across eight schools, which we analyze in full in \cref{sec:eight-schools}. In this model, the population mean $\mu$ and population standard deviation $\tau$ inform group-level means $\lambda_j \sim \mathrm{Normal}(\mu, \tau)$, where $j$ is the group index. The group-level parameters $\lambda_j$ in turn inform the observed data $y_{ji}$, where the index $i$ refers to individual observations in group $j$. The observations additionally depend on a shared observation-level standard deviation $\omega$. This model can be represented by the DAG shown in \cref{fig:graph-pipeline} (left). The nodes $\lambda_j$ and $y_{ji}$ are marked with a dashed box, denoting that the groups are exchangeable.
The factorization of the joint model for this example is given as:
\begin{equation}\label{eqn:joint-factorization-example}
    \begin{aligned}
  &p(\mu, \tau, \omega, \{\lambda_j\}, \{y_{ji}\}) = {} \eqbreak
  p(\mu)\, p(\tau)\,p(\omega) \prod_{j=1}^{J}
    p(\lambda_j \mid \mu, \tau)\, p(y_j \mid \lambda_j, \omega),
    \end{aligned}
\end{equation}
where $y_j$ denotes all observations of group $j$, and the set notation $\{\lambda_j\}$ and $\{y_{ji}\}$ highlights that both the number of groups and the number of observations within groups can vary across datasets.

Two further annotations are required to simulate from the graph. First, each node $v$ carries its conditional distribution $p(v \mid \pa_G(v))$. Second, each node carries a distribution $p(N_v)$ over the number of samples $N_v$ it draws for each combination of its parents' values. Because $N_v$ is drawn anew for every such combination, groups within a dataset may differ in size. Thus, the networks encounter a range of group and dataset sizes during training and can be applied to any size at inference.
An interior node that can draw more than one sample per combination of its parents' values introduces a group index, which is carried by that node and by all of its descendants. We call such a node a \emph{grouping factor}. A grouping factor is \emph{nested} in another if it is a descendant of it, and two grouping factors are \emph{crossed} if neither is nested in the other.
For the two-level model, $N_\lambda$ is the number of groups and $N_y$ is the number of observations within a group. The node $\lambda_j$ carries the group index $j$ it contributes itself, while $y_{ji}$ carries both $j$ and its own observation index $i$. For root nodes, which have no parents, $N_v$ is instead fixed to a common value across all roots, equal to the number of simulated datasets.
\siref{sec:appendix-implementation-two-level} shows a full implementation of the two-level model.

\subsection{Graph inversion}\label{sec:graph-inversion}

We describe graph inversion before graph expansion since inversion motivates why expansion is needed.
\Cref{sec:graphical-simulator} shows that simulating data from $G$ follows a forward direction: nodes are visited in topological order, from marginal prior distributions along conditional dependencies to observed data. When performing parameter inference, the reverse is required: To infer the posterior $p(\theta \mid y)$, a corresponding graph must start from the observed data $y$ and reason back to the latent parameters $\theta$. Following the nomenclature in the literature \citep{Stuhlmueller2013, Webb2018}, we call this process \emph{graph inversion}, and the resulting posterior factorization an \emph{inverse factorization}.

Computing such an inverse factorization is a non-trivial task, as it is not sufficient to simply reverse all edges. To illustrate this, consider $\mu$ and $\tau$ both pointing to $\{\lambda_j\}$: $\mu \rightarrow \{\lambda_j\} \leftarrow \tau$. Two parents with a common child form a \emph{collider}: conditioning on the child makes them dependent, whereas conditioning on a common parent makes its children independent. Reversing the edges yields $\mu \leftarrow \{\lambda_j\} \rightarrow \tau$, which incorrectly asserts independence of $\mu$ and $\tau$ given $\{\lambda_j\}$. In multilevel models, for example, larger population variability in $\tau$ also implies larger uncertainty about the population mean in $\mu$.

To correctly capture this dependency, the inverse graph must additionally feature either an edge from $\mu$ to $\tau$ or from $\tau$ to $\mu$. The former corresponds to first inferring $p(\mu \mid \{\lambda_j\})$ and then $p(\tau \mid \mu, \{\lambda_j\})$; the latter corresponds to first inferring $p(\tau \mid \{\lambda_j\})$ and then $p(\mu \mid \tau, \{\lambda_j\})$. In practice, however, it is often computationally preferable to infer $\mu$ and $\tau$ jointly using a single inference network, learning $p(\mu, \tau \mid \{\lambda_j\})$ directly. This example also demonstrates that, for a given generative model, multiple inverse factorizations may exist.

Assuming $\mu$ and $\tau$ are inferred jointly, this model still has five valid inverse factorizations (\cref{sec:appendix-inverse-factorizations}). Three of these factorizations retain $\{\lambda_j\}$ as a single factor and therefore require all groups to be inferred jointly, breaking amortization. The remaining two factorize the posterior over $\{\lambda_j\}$ into $\prod_{j=1}^{J} p(\lambda_j \mid y_j, \mu, \tau, \omega)$, allowing each group-level parameter to be inferred independently.

\subsubsection{Algorithm}
The example above shows that the choice of inverse factorization has direct consequences for amortization. Algorithms for computing an inverse factorization from the structure of a generative graph have been proposed by Stuhlmüller et al. \citep{Stuhlmueller2013} and Webb et al. \citep{Webb2018}. Our method is agnostic to the chosen inversion algorithm. Here, we build on the algorithm of Stuhlmüller et al., which computes an inverse factorization by ordering nodes and adding them to the inverse graph, with dependencies determined by the original graph structure. Because each node ordering yields one inverse factorization, varying the ordering enumerates the inverse factorizations of a model, from which we then select the most suitable one according to \cref{sec:graph-inversion-expanded}.

Stuhlmüller et al. proposed this algorithm for Bayesian networks of fixed structure, in which the learned inverses amortize across different values observed at the same nodes and serve as proposals within a Monte Carlo sampler. They do not explicitly consider multilevel models. Applied to a multilevel model, their algorithm fixes the number of groups, as each $\lambda_j$ becomes a separate node with its own inverse conditional, valid only for that particular $J$.

We therefore extend the algorithm in two ways: First, we expand exchangeable nodes before inversion (\cref{sec:exchangeable-nodes}), which allows verifying that all groups admit the same inverse conditional. Second, we use the inverse graph not to sample from directly, but to determine the architecture of a set of neural approximators that together enable full amortized posterior inference (\cref{sec:graph-inversion-expanded}).

Given the generative graph $G$, the inverse graph $H$ is constructed as follows: observed nodes are added first, and then latent nodes are processed one by one in a fixed order. When adding a node $v$ to $H$, its parents are set to a minimal subset $S$ of nodes already in $H$ such that, given $S$, knowing the values of any additional node provides no further information about $v$. Formally, $S$ d-separates $v$ from $H \setminus S$ in $G$.

The optimal factorization depends on both the graph structure and the number of groups at each level. 
The cost of graph inversion is independent of the data. Because graph expansion splits each exchangeable node into exactly two instances, the graph being inverted has a fixed size determined by the model specification. Since graph inversion is computationally negligible (on the order of milliseconds compared to minutes or hours for network training), all factorizations of even complex multilevel models can be enumerated, and the most suitable one selected without meaningful overhead (see \cref{sec:graph-inversion-expanded}).

\subsection{Exchangeable nodes and graph expansion}\label{sec:exchangeable-nodes}

Before applying a graph inversion algorithm, any graph containing exchangeable nodes must be expanded to make that exchangeability explicit in its structure. In a generative graph, an exchangeable node stands for an entire population of groups, and their exchangeability is represented only implicitly, through repeated sampling from that node. This suffices for simulation, but the inversion algorithm operates on the graph structure alone: it sees a single node and has no way to know how many groups it represents. D-separation would then force the group-level parameters to be conditioned jointly on the data of all groups, breaking amortization.

To resolve this, each interior node that can draw more than one sample per combination of its parents' values is split into two instances. The inversion algorithm can then verify via d-separation that each instance conditions only on its own data and the global parameters, indicating that all $J$ groups can be treated identically. By symmetry, every pair of instances has the same d-separation relations, so splitting into two instances is sufficient to reveal whether any instance depends on another.

\begin{figure*}[tbhp]
    \centering
    \includegraphics[width=\textwidth]{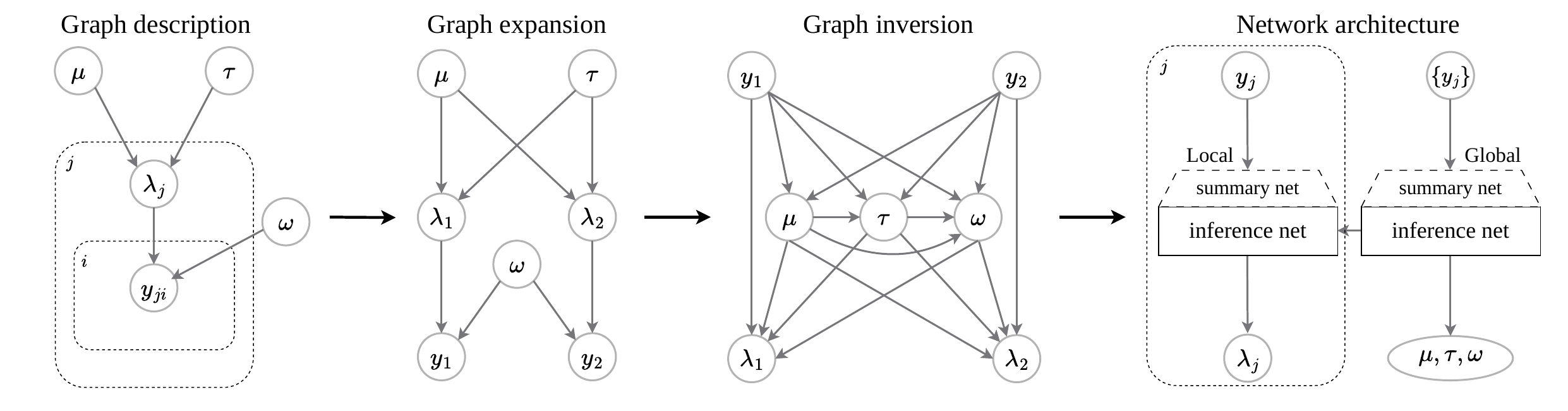}
    \caption{Graph processing pipeline. \textbf{Graph description}: Directed acyclic graph of a two-level model with population mean $\mu$, population standard deviation $\tau$, group-level means $\{\lambda_j\}$, observations $y_{ji}$, and observation-level standard deviation $\omega$. The dashed box denotes that the group-level means $\lambda_j$ and observations $y_{ji}$ are sampled independently. \textbf{Graph expansion}: The interior node $\lambda_j$ is split into two conditionally independent instances $\lambda_1$ and $\lambda_2$, each receiving the same parent nodes $\mu$ and $\tau$. This makes the exchangeability of groups explicit in the graph structure, allowing the inversion to determine that each instance conditions only on its own data and the global parameters. \textbf{Graph inversion}: Inversion of the expanded two-level model using outer-nodes-first ordering $(\mu, \tau, \omega, \lambda_1, \lambda_2)$. In the inverse graph, each $\lambda_j$ conditions only on its own group data $y_j = \{y_{ji} \}$ and the global parameters $\mu$, $\tau$, $\omega$, so the per-group inference becomes amortizable: a single inference network can be used for all $J$ groups. \textbf{Network architecture}: Architecture derived from the inverse graph. Each inference network has its own summary network. The local summary network encodes the observation of each group into a group-level summary, from which the local inference network infers $\lambda_j$. The global summary network pools all observations into a single representation of the dataset, from which the global inference network infers $\mu$, $\tau$, and $\omega$. The dashed box denotes that the local summary and inference networks are applied independently for each group.}
    \label{fig:graph-pipeline}
\end{figure*}

Concretely, the graph expansion procedure is given in \cref{alg:graph-expansion}. \Cref{fig:graph-pipeline} (second panel) shows the expanded version of the generative graph on the left.

\begin{algorithm}[tbhp]
\caption{Graph expansion.}
\label{alg:graph-expansion}
\begin{algorithmic}[1]
\Require generative graph $G$, sample sizes $p(N_v)$
\Ensure expanded graph $G'$
\Statex $\de_{G'}(v)$: descendants of $v$ in $G'$, excluding $v$ itself
\State $G' \gets G$
\State $Q \gets$ nodes of $G$, topologically ordered
\While{$Q \neq \emptyset$}
  \State $v \gets$ pop first element of $Q$
  \If{$v$ interior and $\Pr(N_v > 1) > 0$}
    \State $D \gets$ subgraph of $G'$ induced by $v$ and $\de_{G'}(v)$
    \State $E_{\mathrm{in}} \gets \{u \to w \in G' : u \notin D,\, w \in D\}$
    \State remove $D$ and $E_{\mathrm{in}}$ from $G'$
    \State remove the nodes of $D$ from $Q$
    \For{$k \in \{1, 2\}$}
      \State $D_k \gets$ copy of $D$ with nodes $\{w_k : w \in D\}$
      \State add $D_k$ to $G'$
      \For{each $u \to w \in E_{\mathrm{in}}$}
        \State add edge $u \to w_k$
      \EndFor
      \State append $\de_{G'}(v_k)$ in topological order to $Q$
    \EndFor
  \EndIf
\EndWhile
\end{algorithmic}
\end{algorithm}

\subsection{Graph inversion with expanded graphs}\label{sec:graph-inversion-expanded}

The graph inversion algorithm detailed in \cref{sec:graph-inversion} can now be directly applied to the expanded graph without modifications. Because each exchangeable node now appears as two independent instances, d-separation can determine whether a group-level node depends on data or parameters from outside its own group.

We call a latent node \emph{independently amortizable} when none of its instances condition on another instance of the same node in the inverse graph. All instances of such a node share the same inverse conditional, so a single inference network with shared weights can infer every group independently, applied once per group.
Global parameters, the root nodes of the generative graph, can be merged into a single node and inferred jointly by one inference network.

For the two-level model, inversion with the outer-nodes-first ordering yields the graph in \cref{fig:graph-pipeline} (third panel). Each $\lambda_j$ conditions on its own data $y_j$ and on the global parameters $\mu$, $\tau$, and $\omega$, and is therefore independently amortizable. Merging $(\mu, \tau, \omega)$ reduces the number of required inference networks from four to two: one for the global parameters and one shared across all $J$ groups.

In contrast, the inner-nodes-first ordering $\lambda_1$, $\lambda_2$, $\mu$, $\tau$, $\omega$ yields an inverse graph in which $\lambda_2$ conditions on $\lambda_1$. The group-level parameters are then not independently amortizable, and no single inference network can infer each $\lambda_j$ on its own.

\subsection{Non-independent amortization}\label{sec:non-independent-amortization}

No ordering might yield an inverse factorization in which every grouping factor node is independently amortizable. The typical case is a crossed design, in which observations depend simultaneously on two or more non-nested grouping factors.

Let $v$ and $w$ be exchangeable nodes of two such factors, both parents of an observation node $y$. After graph expansion, $v$ and $w$ appear as instances $v_1, v_2$ and $w_1, w_2$. We write $y_{11}$ for the observations that depend on $v_1$ and $w_1$, and $y_{21}$ for those that depend on $v_2$ and $w_1$.
When constructing the inverse graph, observed nodes enter first since their values are known without inference and are therefore available to condition every node that follows. Conditioning on the observations opens the colliders at $y$, so $v_1$ and $v_2$ remain dependent through the path $v_1 \rightarrow y_{11} \leftarrow w_1 \rightarrow y_{21} \leftarrow v_2$. This path is blocked only by conditioning on $w_1$, that is, only if $w$ enters the inverse graph before $v$.
By symmetry, the same holds for $w_1$ and $w_2$ if $v_1$ was added first. Thus, the instances of whichever factor enters first remain dependent and must be conditioned on one another, whereas the instances of the second factor are d-separated given the first and can be inferred independently.

For crossed factors, a factorization of the posterior can therefore either make one factor independently amortizable or the other, but never both. Inferring the instances of the remaining factor jointly would fix the output dimension, preventing amortization over the number of groups. Thus, we instead infer them autoregressively with a single shared set of weights.

Let $v_1, \ldots, v_K$ denote the instances of a latent node that is not independently amortizable, and let $c_k$ collect the conditions that the inverse graph assigns to $v_k$ apart from the other instances of the same node, with $c = (c_1, \ldots, c_K)$. For any ordering of the instances, the chain rule of conditional probability gives
\begin{equation}\label{eqn:autoregressive-factorization}
    p(v_1, \ldots, v_K \mid c) = \prod_{k=1}^{K} p(v_k \mid v_{1:k-1}, c_k),
\end{equation}
where $v_{1:k-1} = (v_1, \ldots, v_{k-1})$ and $v_{1:0} = \emptyset$. Each factor in \cref{eqn:autoregressive-factorization} is a density over a single instance $v_k$, whose dimension does not depend on $K$. Thus, a single inference network can represent all of them. In contrast, their conditions are not of fixed size: the data and group-level parameters in $c_k$ vary across datasets, and the number of preceding instances $v_{1:k-1}$ grows with $k$.

To rectify this, we compress the conditions of each step into a fixed-size summary. Because the groups are linked through the crossed factor, the observations of every group are informative about $v_k$. A permutation-invariant set network therefore summarizes the observations of all groups, encoded as described in \cref{sec:summary-networks}, together with the preceding $v_{1:k-1}$. Its output then has the same size for every step and every number of groups, so a single inference network can approximate each factor of \cref{eqn:autoregressive-factorization}.

Using this architecture, sequential inference does not force sequential evaluation during training. Because the true values of $v_{1:k-1}$ are provided by the simulator, all $K$ outputs are computed in a single pass. Training is nonetheless more expensive than for an independently amortizable node, since the summary network summarizes a set of $K$ instances at each of the $K$ steps, which scales quadratically with $K$. Only sampling is sequential, and drawing from the approximate posterior therefore requires $K$ passes through the inference network.

The choice of posterior factorization can now be made based on computational efficiency. Both the cost of the summary network during training and the number of passes during sampling grow with the number of instances of the sequentially inferred node. Because all factorizations can be enumerated at negligible cost (\cref{sec:graph-inversion}), we select the one that leaves the fewest instances to be inferred sequentially. This information is available in advance from the sample size distributions $p(N_v)$ provided as part of the model specification. In a crossed design, this amortizes over the grouping factor with the most groups and infers the other sequentially.

\subsection{Summary networks}\label{sec:summary-networks}

Often in amortized Bayesian inference, so-called \emph{summary networks} are trained jointly with the inference networks \citep{Radev2021}. Their purpose is three-fold: First, by learning to compress raw data into informative summary statistics, they provide the inference networks with fixed-width conditions. 
Second, they are used to present the inference networks with a representation that is aligned with the structure of the data. For example, set-based summary networks like DeepSets \citep{Zaheer2017} or SetTransformers \citep{Lee2019} can be used to treat observations as exchangeable, or architectures such as Time Series Transformers \citep{Wen2023} can encode time series.
Third, using a summary network increases computational efficiency if a lower-dimensional representation of the data suffices to inform posterior inference \citep{Marin2012}.

However, in a multilevel model, such summary networks are not applicable without modifications because the observations do not form a single exchangeable set. Observations are exchangeable within a group, and the groups themselves are exchangeable. Consequently, a summary of the observations must reflect this structure. 

Which encodings respect such a structure has been characterized for two basic cases: Hartford et al.~\citep{Hartford2018} show that a linear layer that is equivariant to permutations of the groups of crossed grouping factors must combine each observation with its means over each grouping factor and over all observations. For nested grouping factors, Wang et al.~\citep{Wang2020} show that every linear equivariant layer must pool observations at each level of the hierarchy and then broadcast the result back to the individual observations. The encodings of both crossed and nested grouping factors are thus built by pooling over groups of observations, and the model structure defines over which groups the pooling occurs.

We build this encoding for arbitrary generative graphs. As with the conditions of the inference networks, the generative graph fully defines which grouping factors are nested or crossed, and thus, over which groups the encoding has to be pooled.

The observations can be grouped by any combination of the group indices they carry, with one restriction: a nested grouping factor is only exchangeable within the grouping factors it is nested in. This is because the index of the inner grouping factor refers to a different entity for each index of the outer grouping factor. Concretely, for the breeding bird survey in \cref{sec:bird-survey}, in which squares are nested within regions, the square with index $i$ in region $j$ is not identical to the square with index $i$ in region $k$.

Each inference network receives its own summary network because the parameters it conditions on and the level to which the observations must be pooled differ between networks.
Consider an inference network for a node $v$ that is conditioned on observations and a set of parameters. We summarize these conditions in three steps. First, each observation $y_o$ is concatenated with the parameter values from which it was simulated, restricted to the conditions of $v$. We denote the resulting extended observations as $e_o$. This ensures that every condition enters the summary together with the observations to which it relates. Second, each extended observation $e_o$ is further extended by the log sizes of the groups to which it belongs and is then encoded by a stack of exchangeable layers,
\begin{equation}\label{eqn:exchangeable-layer}
    e_o \leftarrow \sigma\Bigl(W e_o + b + \sum_u W_u \, \bar{e}_{u,o}\Bigr),
\end{equation}
where the sum runs over all groupings $u$, and $\bar{e}_{u,o}$ is the mean of the current encoding over the observations that fall into the same group as $o$ under grouping $u$. That is, each layer updates an observation with the averages over every group to which it belongs. For the breeding bird survey, the count of a square in a given year is thus combined with the averages over the entire dataset, its region, its square, its year, and that region in that year.

Third, the encoded observations are pooled to the level of $v$: for every instance $v_k$, all encoded observations that descend from $v_k$ are averaged, and their log number is appended. This is necessary because averaging discards the number of observations.
Finally, we also directly append the conditioned parameters that are ancestors of $v$. They are already contained in the summary, but appending them passes them to the inference network unchanged and allows conditioning the inference network for groups without observations. The result now has a fixed size, regardless of the number of groups and observations, and serves as the conditions of the inference network.

\subsection{Network architecture}\label{sec:network-architecture}

The network architecture follows directly from the selected inverse graph $H$, corresponding to one factorization of the joint posterior. The observation nodes are the roots of $H$: their values are known, so they require no network. For every other node, the nodes pointing to its instances in $H$ are its conditions.

$H$ also determines how these nodes are grouped into networks. They are sorted into stages based on when their conditions become available: the first stage holds the nodes that condition only on observations, and each subsequent stage holds the nodes whose conditions are inferred at earlier stages. Each node belongs to exactly one stage, and each stage is assigned one inference network, conditioned on the union of its nodes' parents.

Summary and inference networks are trained jointly on simulated pairs $(\theta, y)$, with the parameter conditions taken from the simulator. 
Writing $\phi_n$ for the weights of network $n$, collected in $\phi$, and $\psi_n$ for those of its summary network collected in $\psi$, the networks are trained by minimizing the negative log density of the simulated parameters,
\begin{equation}\label{eqn:architecture-loss}
    \mathcal{L}(\phi, \psi) = -\frac{1}{M} \sum_{m=1}^{M} \sum_{n} \sum_{i}
    \log q_{\phi_n}\bigl(\theta^{(m)}_{n,i} \mid h_{\psi_n}(c^{(m)}_{n,i})\bigr),
\end{equation}
where $h_{\psi_n}(c^{(m)}_{n,i})$ is the fixed-size summary of the conditions of instance $i$, $m$ indexes the $M$ simulated datasets (which may vary in their number of groups and observations), the second sum runs over the inference networks, and the third runs over the instances at which each is applied. The loss $\mathcal{L}(\phi, \psi)$ can be optimized by standard backpropagation, and for a sufficiently large simulation budget, minimizing $\mathcal{L}(\phi, \psi)$ minimizes the expected forward Kullback-Leibler divergence between the approximate and the true posterior \citep{Radev2023}.

During training, the inference networks $n$ can be evaluated in any order since all parameters and observations are known. During sampling, however, the inference networks need to be evaluated in the order in which their conditions become available.
\section{Case studies}\label{sec:case-studies}

We evaluate our method on three applications based on real-world datasets with increasingly complex multilevel structures:

\begin{enumerate}
    \item A two-level model of the ``eight schools'' data of Rubin \citep{Rubin1981}, simple enough that its posterior factorizations can still be derived by hand.
    \item A three-level beta regression model of inter-rater agreement on the ISIC Archive Multi-Annotator Dermoscopic Skin Lesion Segmentation Dataset \citep{Abhishek2025}, with crossed image and annotator effects.
    \item A four-level negative binomial model of yearly bird counts collected in the UK Breeding Bird Survey \citep{Massimino2024}, in which nested region and square effects are crossed with a year effect.
\end{enumerate}

For each case study, we apply a series of model checks. Computational validity, which can be compromised, for example, by incomplete network training, is assessed using simulation-based calibration \citep{Talts2018, Modrak2025}. Inferential accuracy is assessed by posterior predictive checks \citep{Gabry2019} and comparisons to Stan \citep{stan2026} as a gold-standard sampler. Finally, the partial pooling of the group-level parameters $\lambda_j$ toward their hierarchical mean is assessed by a pooling factor \citep{Gelman2006b}.

\subsection{Eight schools}\label{sec:eight-schools}

As a first example, we analyze the eight-schools data of Rubin \cite{Rubin1981}, the running example of \cref{sec:methods}. Eight schools independently conducted randomized experiments to estimate the effect of a coaching program on SAT scores. For each school $j$, the study reports an estimated treatment effect $y_j$ together with its standard error $\sigma_j$, which is treated as known. Because the school-specific effects $\lambda_j$ are assumed to be drawn from a common population, they are modeled hierarchically:

\begin{equation}\label{eqn:eight-schools-model}
    \begin{aligned}
        \mu &\sim \mathrm{Normal}(0, 5), & \tau &\sim \mathrm{Normal}^{+}(0, 20) \\
        \lambda_j &\sim \mathrm{Normal}(\mu, \tau), & y_j &\sim \mathrm{Normal}(\lambda_j, \sigma_j),
    \end{aligned}
\end{equation}

where $\mu$ is the population mean effect and $\tau$ is the between-school standard deviation. Since the standard errors are part of the observed data rather than quantities to be inferred, we simulate them alongside the treatment effects as $\sigma_j \sim \mathrm{Normal}^{+}(15, 5)$, which covers the range of standard errors reported in the original study.

The generative graph of \cref{eqn:eight-schools-model} contains a root node holding $(\mu, \tau)$, an exchangeable node holding $\lambda_j$, and a data node holding $(y_j, \sigma_j)$. Because $\mu$ and $\tau$ are both roots, they carry no group index and are merged into a single node inferred by one network. Of the two inverse factorizations of this graph, our algorithm selects $p(\mu, \tau \mid y) \prod_{j=1}^{J} p(\lambda_j \mid \mu, \tau, y_j)$, which amortizes over the number of schools.

\subsubsection{Network architecture}
The approximator derived from the inverse graph consists of a single summary network, which pools the per-school observations, and two inference networks: one for the merged root node $(\mu, \tau)$ and one for the school effects $\lambda_j$. We use a SetTransformer \citep{Lee2019} with a summary dimension of $16$ as the summary network, and affine coupling flows \citep{Dinh2017} with a depth of $12$ as the inference networks.

\subsubsection{Model training} 
We generate training data by ancestral sampling from \cref{eqn:eight-schools-model}, using a simulation budget of $M = 100\,000$ datasets. The number of schools is drawn uniformly from $[1, 100]$ for each simulated dataset, inference on the original data is then performed at $J=8$. The parameter $\tau$ and the observed standard errors $\sigma_j$ are log-transformed before training. The networks are trained jointly for 100 epochs with a batch size of 4096 using the Adam optimizer \citep{Kingma2015} with a cosine decay schedule from an initial learning rate of $10^{-3}$.

\subsubsection{Results}
\begin{figure*}
    \centering
    \includegraphics[width=\textwidth]{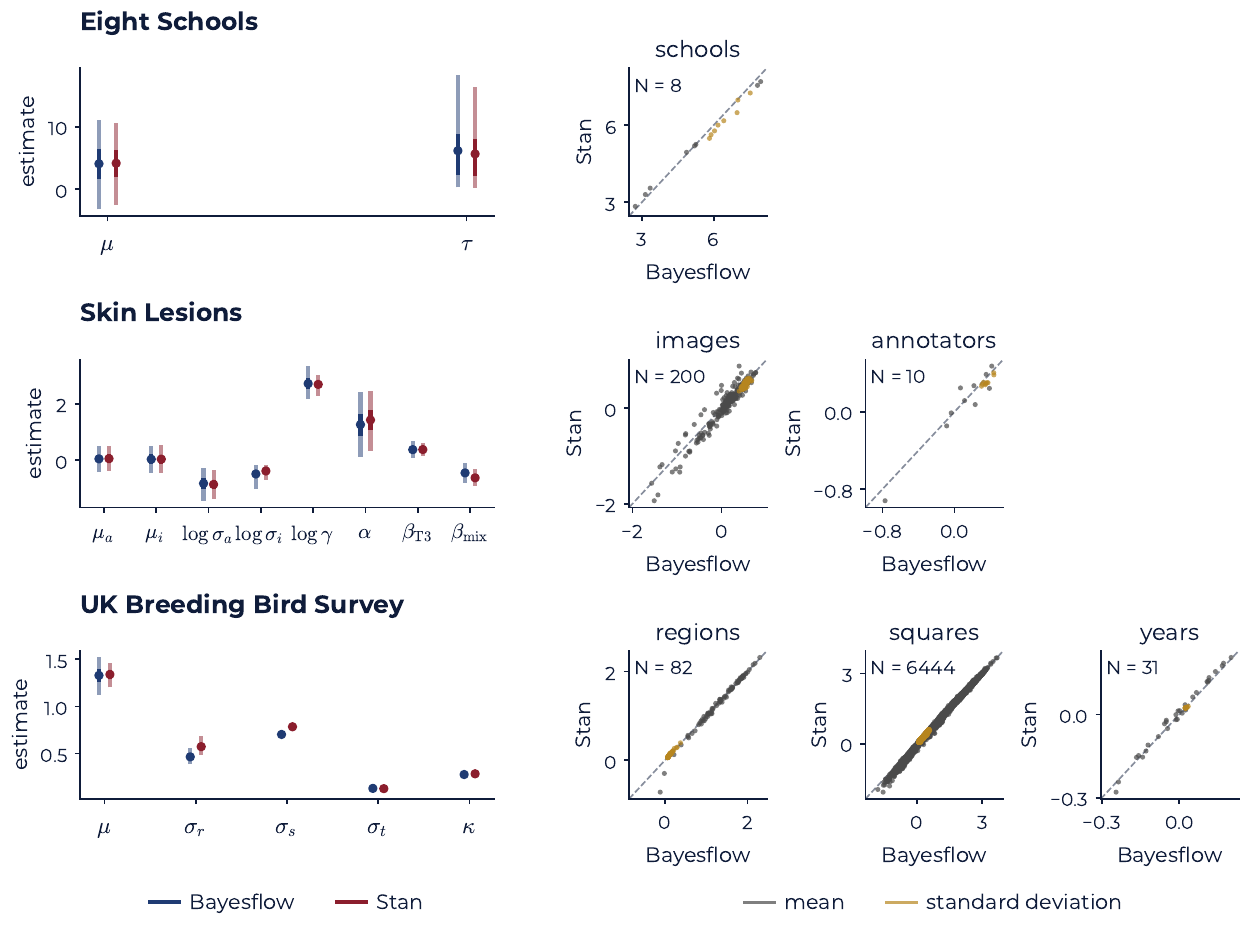}
    \caption{Comparison of our method against Stan on the three case studies. \textbf{Left:} marginal posterior intervals of the global parameters, showing the posterior mean (point) and 90\% credible intervals (line) under our approximator (blue) and Stan (red). \textbf{Right}: The posterior mean (grey) and posterior standard deviation (gold) of each group-level parameter under our method, plotted against the corresponding Stan estimate. The dashed line marks exact agreement and $N$ is the number of groups.}
    \label{fig:posterior-summary}
\end{figure*}

\Cref{fig:posterior-summary} (top row) compares marginal posterior intervals for $\mu$, $\tau$, and the school-specific effects $\lambda_j$ on the original data of Rubin \citep{Rubin1981} against a reference implementation in Stan \citep{stan2026}. The estimates agree closely, showing that the architecture derived from the inverse graph recovers the correct posterior. Network training takes about a minute on a standard desktop computer, whereas drawing 4000 independent posterior samples takes about $0.04\,\mathrm{s}$, approximately as long as Stan requires to reach an effective sample size of 4000 for this model.

\subsection{Skin Lesions} \label{sec:skin-lesions}

To demonstrate our method on a three-level model with a crossed design, we analyze the ISIC Archive Multi-Annotator Dermoscopic Skin Lesion Segmentation Dataset \citep{Abhishek2025}. The dataset consists of dermoscopic images of skin lesions, each of which was segmented by several expert annotators who separated the region of the lesion from the surrounding skin. Because each image was annotated by multiple annotators and each annotator segmented multiple images, image and annotator effects are crossed rather than nested. 

We model how strongly two annotators agree on a given image. For each image with at least two annotators, we compute the Dice similarity coefficient between the masks of every pair of annotators, yielding one response in $(0, 1)$ per image and annotator pair. Because the response is a proportion, we model it using a beta regression with crossed image and annotator effects:
\begin{equation}\label{eqn:skin-lesion-model}
    \begin{gathered}
    \begin{alignedat}{2}
        \alpha &\sim \mathrm{Normal}(1, 1), &\beta_{\mathrm{T3}}, \beta_{\mathrm{mix}} &\sim \mathrm{Normal}(0, 0.5) \\
        \gamma &\sim \mathrm{LogNormal}(\log 15, 0.5), &\mu_I, \mu_A &\sim \mathrm{Normal}(0, 0.25) \\
        \sigma_I &\sim \mathrm{Normal}^{+}(0, 0.5), &\sigma_A &\sim \mathrm{Normal}^{+}(0, 0.3) \\
        u_j &\sim \mathrm{Normal}(\mu_I, \sigma_I), &v_k &\sim \mathrm{Normal}(\mu_A, \sigma_A)
    \end{alignedat} \\
    \begin{aligned}
        \theta_{jkl} &= \mathrm{logit}^{-1}\!(\alpha + u_j + v_k + v_l + \beta_{\mathrm{T3}} c_{kl} + \beta_{\mathrm{mix}} m_{kl}) \\
        y_{jkl} &\sim \mathrm{Beta}(\theta_{jkl}\gamma,\; (1 - \theta_{jkl})\gamma)
    \end{aligned}
    \end{gathered}
\end{equation}
where $y_{jkl}$ is the observed agreement between annotators $k$ and $l$ on image $j$, and $\theta_{jkl}$ is its expected value. The beta likelihood is parameterized by $\theta_{jkl}$ and the precision $\gamma$, so a larger $\gamma$ concentrates responses more tightly around $\theta_{jkl}$. The intercept $\alpha$ is the population-level agreement on the logit scale. The image effect $u_j$ and the annotator effect $v_k$ are drawn around means $\mu_I$ and $\mu_A$ with standard deviations $\sigma_I$ and $\sigma_A$. Because each response compares two annotators, the agreement between annotators $k$ and $l$ on image $j$ depends on both $v_k$ and $v_l$.

The dataset records which tool the annotators used to produce a segmentation: a manually traced polygon or a fully automated algorithm. Two covariates carry that information. The first, $c_{kl} \in \{0, 1, 2\}$, counts how many of the two masks were produced automatically, so $\beta_{\mathrm{T3}}$ is the change in agreement contributed by each automated mask. The second, $m_{kl}$, indicates that the two annotators used different tools, so $\beta_{\mathrm{mix}}$ captures the agreement lost because the boundaries were drawn by different procedures rather than because either was automated.

The generative graph of \cref{eqn:skin-lesion-model} contains two root nodes, one holding the hyperparameters $\eta = (\mu_I, \sigma_I, \mu_A, \sigma_A)$ and the other holding the shared parameters $\xi = (\alpha, \gamma, \beta_{\mathrm{T3}}, \beta_{\mathrm{mix}})$, two exchangeable nodes holding the image effects $u_j$ and the annotator effects $v_k$, respectively, and a data node holding $y_{jkl}, c_{kl}, m_{kl}$. Because both exchangeable nodes are parents of the same observation node, no factorization makes both of them independently amortizable (\cref{sec:non-independent-amortization}). Our algorithm  selects the factorization that leaves the factor with the fewest instances to be inferred autoregressively,
\begin{equation}\label{eqn:skin-lesion-factorization}
    \begin{aligned}
        &p(\eta, \xi, \{v_k\}, \{u_j\} \mid y, c, m) = p(\eta, \xi \mid y, c, m) \\
        &\quad \times \prod_{k=1}^{K} p(v_k \mid v_{1:k-1}, \eta, \xi, y, c, m) \eqbreak
        \times \prod_{j=1}^{J} p(u_j \mid \eta, \xi, \{v_k\}, y_j, c, m),
    \end{aligned}
\end{equation}
which amortizes over the $J=2394$ images and infers the $K=15$ annotator effects autoregressively.

\subsubsection{Network architecture}
Three affine coupling flows \citep{Dinh2017} with a depth of $10$ are used as inference networks: one for the merged root nodes $\eta$ and $\xi$, and one for each of the two sets of group-level effects. Their conditions are reduced by five summary networks with a summary dimension of $128$: a two-step chain (\cref{sec:summary-networks}) over the observation tensor, a per-image summary of the same tensor, a pooling of the annotator effects, and an encoder for the preceding draws $v_{1:k-1}$.

\subsubsection{Model training}
We generate training data by ancestral sampling from \cref{eqn:skin-lesion-model}, using a simulation budget of $M=200\,000$ datasets. We vary the number of annotators $K$ uniformly in $K \in [1, 25]$ and the number of images $J$ uniformly in $J \in [1, 2500]$. In the observed data, not every image is annotated by every annotator. We replicate this sparsity pattern by randomly masking the observed data with a binary indicator of whether the annotator segmented that image. Both $\gamma$ and the two group-level standard deviations are log-transformed before training. The networks are trained jointly for 200 epochs with a batch size of 64 using the Adam optimizer \citep{Kingma2015} with a cosine decay schedule from an initial learning rate of $10^{-3}$.

\subsubsection{Results}
\Cref{fig:posterior-summary} (center row) compares the marginal posterior intervals of the global parameters with those obtained by Stan \citep{stan2026}, together with the posterior means and standard deviations of all image and annotator effects. The estimates obtained by our model agree well with the gold-standard results.
Network training takes about one and a half hours on a desktop GPU, whereas drawing 4000 independent posterior samples takes about $2\,\mathrm{s}$, compared to about $13\,\mathrm{min}$ for Stan to reach an effective sample size of 4000. This is a type of model that can already benefit from amortized inference; for example, if many model refits are required for cross validation. With these timings, the training cost is recovered after about seven model refits.

\subsection{UK Breeding Bird Survey} \label{sec:bird-survey}

As an example of a four-level model, we analyze the UK breeding bird survey \citep{Massimino2024}. For this data, volunteers walk two transects across a  $1\,\mathrm{km}$ grid square twice per breeding season and record every bird they encounter, so the survey yields two counts per square per year. Squares are nested within survey regions, while the year of the survey applies to every square. Region and square effects are therefore crossed with a year effect.

We analyze the counts of the European robin. In accordance with the survey guidelines, we take the larger of the two visit counts for each square and year, and we treat a square that was surveyed without recording a robin as having a count of zero. We exclude the $628$ of $7\,072$ surveyed squares in which no robin was recorded in any year. The resulting dataset comprises $N = 89\,865$ observations on $6\,444$ squares in $82$ regions over the $31$ years from 1994 to 2024. That is, $45\%$ of all square-year combinations are observed: regions contain between $4$ and $246$ squares, and a square was surveyed in between $1$ and $31$ years. A small number of squares cover twice the standard area, which we account for by an offset $o_{rst}= \log 2$ on the log scale. Because the counts are overdispersed, we model them with a three-level negative binomial regression,
\begin{equation}\label{eqn:bird-survey-model}
    \begin{gathered}
    \sigma_r, \sigma_s, \sigma_t, \kappa \sim \mathrm{Normal}^{+}(0, 1) \\
    \begin{alignedat}{2}
        \mu &\sim \mathrm{Normal}(0, 5), &\qquad \beta_t &\sim \mathrm{Normal}(0, \sigma_t) \\
        \alpha_r &\sim \mathrm{Normal}(\mu, \sigma_r), & \alpha_{rs} &\sim \mathrm{Normal}(\alpha_r, \sigma_s)
    \end{alignedat} \\
    y_{rst} \sim \mathrm{NegBinomial}\big(\exp(\alpha_{rs} + \beta_t + o_{rst}),\; \kappa\big),
    \end{gathered}
\end{equation}
where $\mathrm{NegBinomial}(m, \kappa)$ denotes the negative binomial distribution with mean $m$ and variance $m + \kappa^2 m^2$. Here, $\mu$ is the mean log-intensity, $\alpha_r$ is the effect of region $r$, $\alpha_{rs}$ is the effect of square $s$ within region $r$, and $\beta_t$ is the effect of year $t$. The year effects are centered at zero, so that $\mu$ carries the overall level and $\beta_t$ describes the deviation of each year from it.

The generative graph of \cref{eqn:bird-survey-model} contains a root node holding $(\mu, \sigma_r, \sigma_s, \sigma_t, \kappa)$, three exchangeable nodes holding the region, square, and year effects, respectively, and a data node holding $(y_{rst}, o_{rst})$. The square node is a descendant of the region node, whereas the year node shares only the root node with them, and both reach the same observations. Writing $\xi=(\mu, \sigma_r, \sigma_s, \sigma_t, \kappa)$ for the global parameters, our algorithm selects the factorization
\begin{equation}\label{eqn:bird-survey-factorization}
    \begin{aligned}
        &p(\xi, \{\beta_t\}, \{\alpha_r\}, \{\alpha_{rs}\} \mid y) = p(\xi \mid y) \prod_{t=1}^{T} p(\beta_t \mid \beta_{1:t-1}, \xi, y) \\
        &\quad \times \prod_{r=1}^{R} p(\alpha_r \mid \xi, \{\beta_t\}, y) \prod_{r,s} p(\alpha_{rs} \mid \xi, \alpha_r, \{\beta_t\}, y).
    \end{aligned}
\end{equation}
The factorization amortizes over the $R=82$ regions and the $6\,444$ squares, leaving only the $T=31$ year effects to be estimated autoregressively by the mechanism of \cref{sec:non-independent-amortization}. Sampling therefore requires $T$ passes through the sequential inference network, in addition to one pass for the global, region, and square effects, respectively.

\subsubsection{Network architecture}
The approximator derived from \cref{eqn:bird-survey-factorization} consists of four affine coupling flows \citep{Dinh2017}: one for the global parameters $\xi$ with a depth of $8$ and hidden layers of width $256$, and one each for the year, region, and square effects with a depth of $6$ and hidden layers of width $128$. Their conditions are reduced by eight DeepSets \citep{Zaheer2017}.

\subsubsection{Model training}
We generate training data by ancestral sampling from \cref{eqn:bird-survey-model}, using a simulation budget of $M = 1\,000\,000$ datasets. Each dataset has between $1$ and $100$ regions with up to $250$ squares each, observed over $T=31$ years. Counts are passed to the networks as $\log(1 + y_{rst})$ together with the offset.
The networks are trained jointly for 100 epochs with a batch size of 2 and gradients accumulated over 4 batches, using the Adam optimizer \citep{Kingma2015} with a cosine decay schedule from an initial learning rate of $10^{-3}$ to $10^{-5}$ and a global gradient norm clip of $1$. 

Counts are passed to the networks as $\log(1 + y_{rst})$ together with the offset. The networks are trained jointly for 200 epochs with a batch size of 512 using the Adam optimizer \citep{Kingma2015} with a cosine decay schedule from an initial learning rate of $10^{-3}$.

\subsubsection{Results}
\Cref{fig:posterior-summary} (bottom row) compares marginal posterior estimates of the neural approximator with those obtained by Stan. Generally, the posterior estimates agree well with the gold-standard sampler results. One exception is $\sigma_r$ and $\sigma_s$, which are slightly too low.
Network training takes about $12$ hours on a desktop GPU, whereas sampling an effective sample size of 4000 draws takes about $4\,\mathrm{s}$, compared with about $35\,\mathrm{min}$ for Stan.
\section{Discussion} \label{sec:discussion}

We have developed a method that derives neural architectures for amortized Bayesian inference on arbitrary multilevel models automatically from the model specification. Graph expansion and graph inversion reduce the choice of posterior factorization to a question of d-separation, and the resulting inverse graph determines the full architecture. Every conditional independence and exchangeability assumption of the generative model is retained. In crossed designs, where no factorization makes every grouping factor independently amortizable, autoregressive inference keeps the approximator amortized over the number of groups.

Across three case studies, the derived approximators closely matched Stan on models with more than 6,500 parameters, while reducing inference to a near-instant forward pass. Although these models have tractable likelihoods to allow the comparison with Stan, the method does not require one and thus also applies to purely simulation-based multilevel models.

However, amortization over the number of groups typically holds only within the range seen during training, so models with many groups require training on many groups. Combining our method with compositional score matching \citep{Arruda2026b} could relax this requirement. Further, as any method trained on simulations, the approximators may become inaccurate under model misspecification \citep{Schmitt2023}. Methods such as self-consistency losses \citep{Mishra2026} improve robustness in such out-of-simulation settings and, since they modify the training objective rather than the architecture, can in principle be combined with out approach.

Together, these results establish graph-based derivation as a principled and general route to amortized inference of multilevel models of arbitrary structure.

\section*{Acknowledgments}
This work was supported by Deutsche Forschungsgemeinschaft (DFG, German Research Foundation) Projects 508399956, 528702768, and 569534706. Paul-Christian Bürkner acknowledges support from DFG Collaborative Research Center 391 (Spatio-Temporal Statistics for the Transition of Energy and Transport) – 520388526. Stefan T. Radev was funded by the National Science Foundation under Grant No. 2448380.

\bibliographystyle{unsrtnat}
\bibliography{bibliography}

\clearpage
\appendix
\crefalias{section}{appendix}   
\section{Enumerating inverse factorizations}\label{sec:appendix-inverse-factorizations}

\begin{align*}
(1)\quad p(\mu, \tau, \omega, \{\lambda_j\} \mid \{y_{ij}\}) &=
  p(\mu, \tau \mid \{y_{ij}\})\, p(\{\lambda_j\} \mid \{y_{ij}\}, \mu, \tau)\,
  p(\omega \mid \{y_{ij}\}, \{\lambda_j\}) \\
(2)\quad p(\mu, \tau, \omega, \{\lambda_j\} \mid \{y_{ij}\}) &=
  p(\mu, \tau \mid \{y_{ij}\}) \prod_{j=1}^{J} p(\lambda_j \mid y_j, \mu, \tau,
  \omega)\; p(\omega \mid \mu, \tau, \{y_{ij}\}) \\
(3)\quad p(\mu, \tau, \omega, \{\lambda_j\} \mid \{y_{ij}\}) &=
  p(\mu, \tau \mid \{\lambda_j\})\, p(\{\lambda_j\} \mid \{y_{ij}\}, \omega)\,
  p(\omega \mid \{y_{ij}\}) \\
(4)\quad p(\mu, \tau, \omega, \{\lambda_j\} \mid \{y_{ij}\}) &=
  p(\mu, \tau \mid \{y_{ij}\}, \omega) \prod_{j=1}^{J} p(\lambda_j \mid y_j, \mu,
  \tau, \omega)\, p(\omega \mid \{y_{ij}\}) \\
(5)\quad p(\mu, \tau, \omega, \{\lambda_j\} \mid \{y_{ij}\}) &=
  p(\mu, \tau \mid \{\lambda_j\})\, p(\{\lambda_j\} \mid \{y_{ij}\})\,
  p(\omega \mid \{y_{ij}\}, \{\lambda_j\})
\end{align*}

Factorizations $(1)$, $(3)$, and $(5)$ infer $\{\lambda_j\}$ jointly, breaking amortization over the number of groups. In contrast, factorizations $(2)$ and $(4)$ estimate each group-level effect independently instead.

\section{Full implementation of a two-level hierarchical model} \label{sec:appendix-implementation-two-level}

\begin{figure*}[tbhp]
  \centering
  \includegraphics[width=\textwidth]{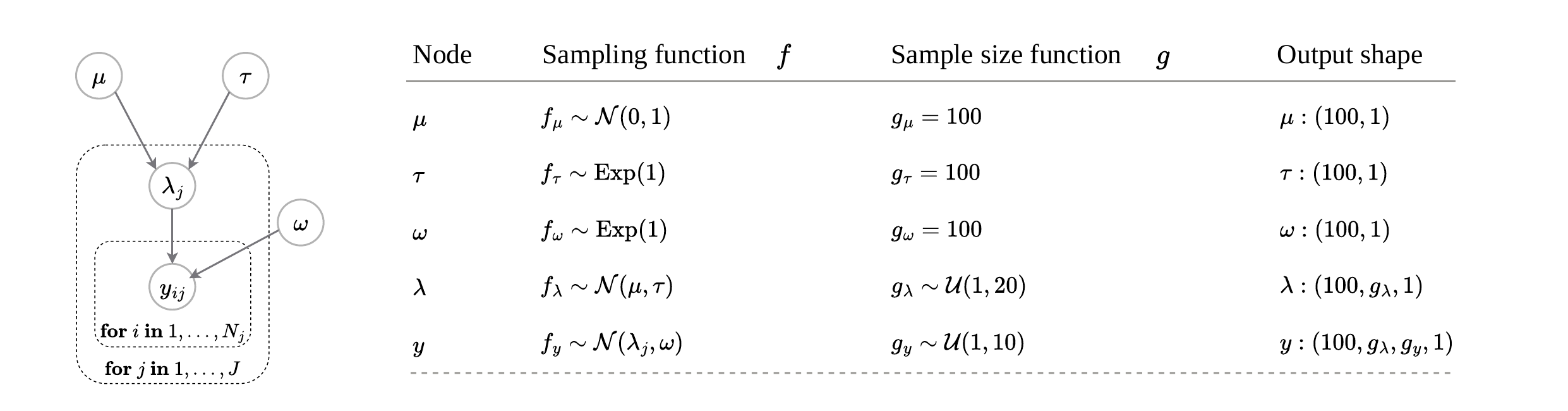}
  \caption{Annotated two-level model graph (left) with its function specification (right). Each node is annotated with a sampling function $f$ and a sample size function $g$. The root nodes $\mu$, $\tau$ and $\omega$ have fixed sample sizes of 100. $\lambda_j$ draws a sample size uniformly from $[1, 20]$ and observations draw a sample size uniformly from $[1, 10]$. The right columns the output shape for each node, where the trailing dimension is the data dimension (always 1 for scalar parameters).}
  \label{fig:twolevel-annotated}
\end{figure*}

To make this point concrete, the table in \cref{fig:twolevel-annotated} shows an example annotation for the two-level model on the left-hand side. The sampling functions $f$ and sample size functions $g$ are provided by the user. To allow for arbitrary nesting, the generated samples are internally stored in long format along with their group index. For each internal node, creating new draws corresponds to creating a full join of all parent nodes using the index columns as keys. For each row in this dataframe, we then generate a sample size $s$ by running the respective sample size function, use the row values as input for the sampling function and execute this function $s$ times with constant inputs. The generated draws are assigned a new index and appended to a new dataframe, along with the indices of the inputs.

\end{document}